\documentclass[runningheads]{llncs}
\usepackage[T1]{fontenc}
\usepackage{graphicx,verbatim}
\usepackage{amsmath}
\usepackage{amssymb}
\usepackage{url}
\usepackage{multirow}
\usepackage{bbding}
\usepackage[colorlinks,linkcolor=red, anchorcolor=blue, citecolor=blue, urlcolor=magenta]{hyperref}
\begin{document}
\title{Delving into Out-of-Distribution Detection with Medical Vision-Language Models}
\titlerunning{OOD detection with Medical VLMs}
\begin{comment}  %% Removed for anonymized MICCAI 2025 submission
\author{First Author\inst{1}\orcidID{0000-1111-2222-3333} \and
Second Author\inst{2,3}\orcidID{1111-2222-3333-4444} \and
Third Author\inst{3}\orcidID{2222--3333-4444-5555}}
%
\authorrunning{F. Author et al.}
\institute{Princeton University, Princeton NJ 08544, USA \and
Springer Heidelberg, Tiergartenstr. 17, 69121 Heidelberg, Germany
\email{lncs@springer.com}\\
\url{http://www.springer.com/gp/computer-science/lncs} \and
ABC Institute, Rupert-Karls-University Heidelberg, Heidelberg, Germany\\
\email{\{abc,lncs\}@uni-heidelberg.de}}

\end{comment}

\author{Lie Ju\inst{1,2,3,4}, Sijin Zhou\inst{1,4}, Yukun Zhou\inst{2,3}, Huimin Lu\inst{5}, \\ Zhuoting Zhu\inst{6},   Pearse A. Keane\inst{2,3}, \and Zongyuan Ge\inst{1,4,}\Envelope}  %% Added for anonymized MICCAI 2025 submission
\authorrunning{L. Ju et al.}

% \institute{Monash Medical AI Group, Monash University, Australia \and  Moorfields Eye Hospital, United Kingdom \and Institute of Ophthalmology, University College London, United Kingdom
%   \and
%  School of Automation, Southeast University, China \and
%  Centre for Eye Research Australia, Melbourne University, Australia
%  \\
%     \email{\{Lie.Ju1, Zongyuan.Ge\}@monash.edu}}

\institute{Monash University, Australia \and  Moorfields Eye Hospital, United Kingdom \and University College London, United Kingdom
  \and
  Airdoc Technology Inc, China \and
  Southeast University, China \and
 Melbourne University, Australia
 \\
    \email{\{Lie.Ju1, Zongyuan.Ge\}@monash.edu}}

\maketitle              % typeset the header of the contribution
\begin{abstract}
Recent advances in medical vision-language models (VLMs) demonstrate impressive performance in image classification tasks, driven by their strong zero-shot generalization capabilities. However, given the high variability and complexity inherent in medical imaging data, the ability of these models to detect out-of-distribution (OOD) data in this domain remains underexplored. In this work, we conduct the first systematic investigation into the OOD detection potential of medical VLMs. We evaluate state-of-the-art VLM-based OOD detection methods across a diverse set of medical VLMs, including both general and domain-specific purposes. To accurately reflect real-world challenges, we introduce a cross-modality evaluation pipeline for benchmarking full-spectrum OOD detection, rigorously assessing model robustness against both semantic shifts and covariate shifts. Furthermore, we propose a novel hierarchical prompt-based method that significantly enhances OOD detection performance. Extensive experiments are conducted to validate the effectiveness of our approach. The codes are available at \url{https://github.com/PyJulie/Medical-VLMs-OOD-Detection}.

% Authors must provide keywords and are not allowed to remove this Keyword section.

\end{abstract}
\section{Introduction}
Recent advances in vision-language models (VLMs), exemplified by CLIP~\cite{radford2021learning}, have significantly advanced image recognition through remarkable generalization capabilities, particularly in zero-shot transfer learning. This success has catalyzed a growing interest in the development of medical VLMs, ranging from general-purpose architectures~\cite{zhang2023biomedclip,lin2023pmc,khattak2024unimed} to domain-specialized experts (e.g., ophthalmology~\cite{wu2024mm,wang2024common,silva2025foundation}). Although these models demonstrate high accuracy on in-distribution (ID) samples with textual descriptions, the ability of medical VLMs to distinguish out-of-distribution (OOD) samples remains unclear, especially considering the potential for overconfident predictions on these OOD samples. In this context, failing to address or assess OOD samples inappropriately can lead to severe outcomes such as misdiagnosis, which could endanger individuals in deployed medical systems~\cite{hong2024out}.

Although many OOD detection methods for image classification have achieved remarkable progress in recent years~\cite{lee2018simple,liang2018enhancing}, conventional vision encoder-only models typically encoded the categories into one-hot vector, leaving the semantic information encapsulated in texts largely unexploited. To utilize the natural advantages of VLMs and address the relevant challenges, some VLM-based methods have been developed for natural images OOD detection, primarily operating under zero-shot~\cite{ming2022delving,miyai2023zero} or few-shot paradigms~\cite{ming2024does,miyai2023locoop}. Zero-shot OOD detection does not require additional training with in-distribution data and typically relies on post-processing techniques. In contrast, few-shot OOD detection involves learning from ID data during both the training and inference phases. To the best of our knowledge, no research has yet explored generalized OOD detection based on medical VLMs, and few studies on medical VLMs have included an analysis of OOD detection capabilities in their experiments. Importantly, it remains unvalidated whether these methods designed for natural image OOD detection are applicable to medical images.

\begin{figure}[t]
    \centering
    \includegraphics[width=12cm]{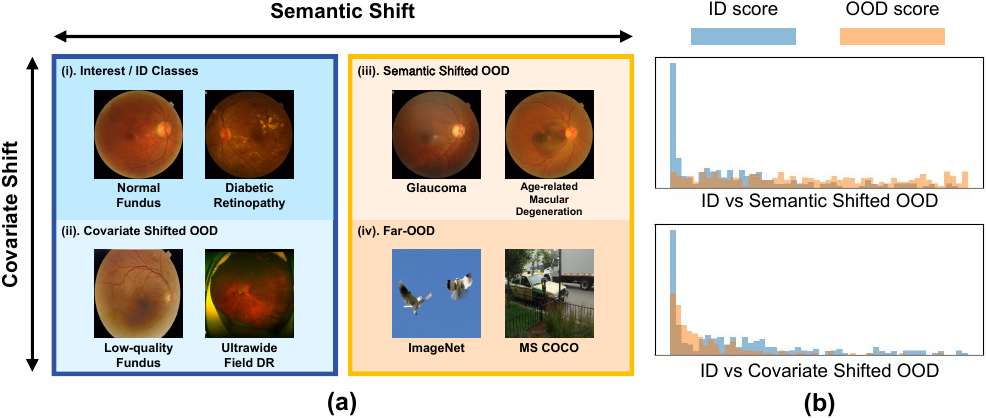}
    \caption{(a) Problem illustration. (i) ID classes with textual descriptions seen by CLIP-like models are defined as ID classes (e.g., diabetic retinopathy); (ii) Covariate shifted OOD data sharing semantic relevance with ID classes but exhibiting covariate shifts, such as low image quality or differences in imaging devices (e.g., ultrawide-field fundus imaging). (iii) \& (iv) OOD with irrelevant concerns of semantics. (b) A simple baseline experiment demonstrates that advanced OOD detection techniques (e.g., MCM~\cite{ming2022delving}) tend to fail on covariate-shifted OOD scenarios.  }
    \label{fig:problem definition}
\end{figure}

Another major challenge in medical OOD detection is understanding how different types of shifts lead to the exclusion of OOD samples from ID data~\cite{hong2024out,noda2025benchmark}. Previous OOD detection benchmarks primarily focus on identifying outliers with \textbf{semantic shifts}, where the semantic labels of OOD samples do not overlap with those of ID samples. For example, predicting natural images using an ophthalmology-specific VLM. Recent OOD detection research has shifted its focus to a more challenging and realistic problem setting: \textbf{covariate shifts}. Unlike semantic shifts, covariate shifts do not alter the target categories, meaning OOD samples share the same semantic labels with ID samples but however differ in imaging attributes such as \textit{imaging modalities}, \textit{imaging quality}, or \textit{population distributions}~\cite{hong2024out}. A typical example is diagnosing lung opacity using a model trained on X-ray images but CT images are incorrectly input. We summarized two types of OOD shifts in Fig.~\ref{fig:problem definition}, which also reveals that advanced OOD detection techniques often struggle in such scenarios. The evaluation of OOD detection performance across both semantic shifts and covariate shifts is defined as \textbf{full-spectrum OOD detection}~\cite{yang2023full}.

In this work, we first evaluate state-of-the-art VLM-based OOD detection methods on a set of medical VLMs, including both general-purpose and domain-specific models. Subsequently, we propose a hierarchical prompt-based method that works on both zero-shot inference phase and retraining phase with few-shot fine-tuning. We also define three dataset evaluation pipelines to establish a novel benchmark which simulates challenging conditions for real-world applications.

Based on the above perspectives, the main contributions of this paper are:

\begin{enumerate}
    \item We present the first systematic evaluation of generalized OOD detection capabilities in medical VLMs. By integrating state-of-the-art OOD detection methods and assessing both general-purpose and domain-specific medical VLMs, we establish a novel benchmark that rigorously addresses the unique challenges of medical imaging, including semantic and covariate shifts.
    \item To bridge the gap between synthetic evaluations and real-world clinical scenarios, we design three benchmark datasets that simulate full-spectrum OOD detection challenges. These datasets encompass diverse imaging modalities, anatomical regions, and distribution shifts, aiming to build holistic evaluation under conditions mirroring clinical deployment.
    \item We propose a novel hierarchical prompt framework that leverages structured medical semantics to improve OOD separability. Validated through extensive experiments, our method demonstrates consistent improvements over existing baselines in challenging scenarios.
\end{enumerate}

\section{Preliminaries}

\subsection{Contrastive vision-language models}
Recent large vision-language models have shown great potential for various computer vision tasks. In this paper, we focus on CLIP-like models which jointly model the visual and textual data using a dual encoder architecture with one visual encoder \(\mathbf{v}_i = f_\theta(\cdot) \in \mathbb{R}^d\) and one text encoder $\mathbf{t}_j = g_\phi(\cdot) \in \mathbb{R}^d$, where $\theta$ and $\phi$ are the corresponding parameters. 
Formally, for an image $x_{i}$ out of all images $X$, the cosine similarity with the specific category $y_{j}$ out of all candidate prompts $Y$ can be calculated as:
\begin{equation}
    s_{i,j} = \frac{\mathbf{v}\cdot \mathbf{t}}{\Vert \mathbf{v}\Vert \cdot \Vert \mathbf{t} \Vert}.
\end{equation}
Commonly, we select the text with the highest similarity as the prediction result:
\begin{equation}
    \hat{i} = \text{arg} \underset{i=1, ..., M}{\text{max}} \mathbf{s_{i}}.
\end{equation}

\subsection{OOD Detection with CLIP}
A notable advantage of CLIP is its ability, known as zero-shot transfer, to make predictions on any potential categories given a set of candidate prompts. However, this also raises the risk of generating blind predictions for OOD samples that do not belong to any of the provided categories. Given an input $x_{i}$, A baseline for OOD detection with CLIP can be formulated as:
\begin{equation}
    G(x_{i}; f,g) = \left\{
	\begin{aligned}
		1 & , & S(x_{i}; f,g) \ge \lambda \\
		0 & , & S(x_{i}; f,g) < \lambda
	\end{aligned} \quad ,
	\right.
\end{equation}
where $S(\cdot)$ is a scoring function to measure the possibility that the input sample $x_{i}$ is an OOD sample and $\lambda$ is set manually as the threshold. A simple scoring function can be directly based on the maximum logit score, which is:
\begin{equation}
    S_{MS} = -\text{max} \; \mathbf{s_{i}}.
\end{equation}
Under this formulation, the prediction probability with a lower confidence should have a higher probability to be an OOD sample.

\subsection{Maximum Concept Matching as Scoring Function}
Maximum Concept Matching (MCM)~\cite{ming2022delving} is a state-of-the-art zero-shot OOD detection method that calculates the OOD confidence after softmax function with proper temperature scaling $\tau$. Given the cosine similarity $\mathbf{s_{i}}$, we have:
\begin{equation}
    S_{\text{MCM}} = -\text{max} \frac{e^{s_{i,j}/\tau}}{\sum_{j=1}^{M} e^{s_{i,j}/\tau}},
\end{equation}
where the temperature value $\tau$ is depended on the downstream datasets. MCM stated that softmax with temperature scaling improves the separability between ID and OOD samples. MCM also suggested that naive maximum softmax probability (MSP) without temperature is suboptimal for zero-shot OOD detection. In this paper, we introduce MCM as a strong baseline for OOD scoring distribution function as its simple nature without the requirements of re-training or complex hyper-parameters tuning. Unless specified, this work uses MCM as the OOD scoring function for all evaluated comparison methods.

\section{Methodologies}
\begin{figure}
    \centering
    \includegraphics[width=12cm]{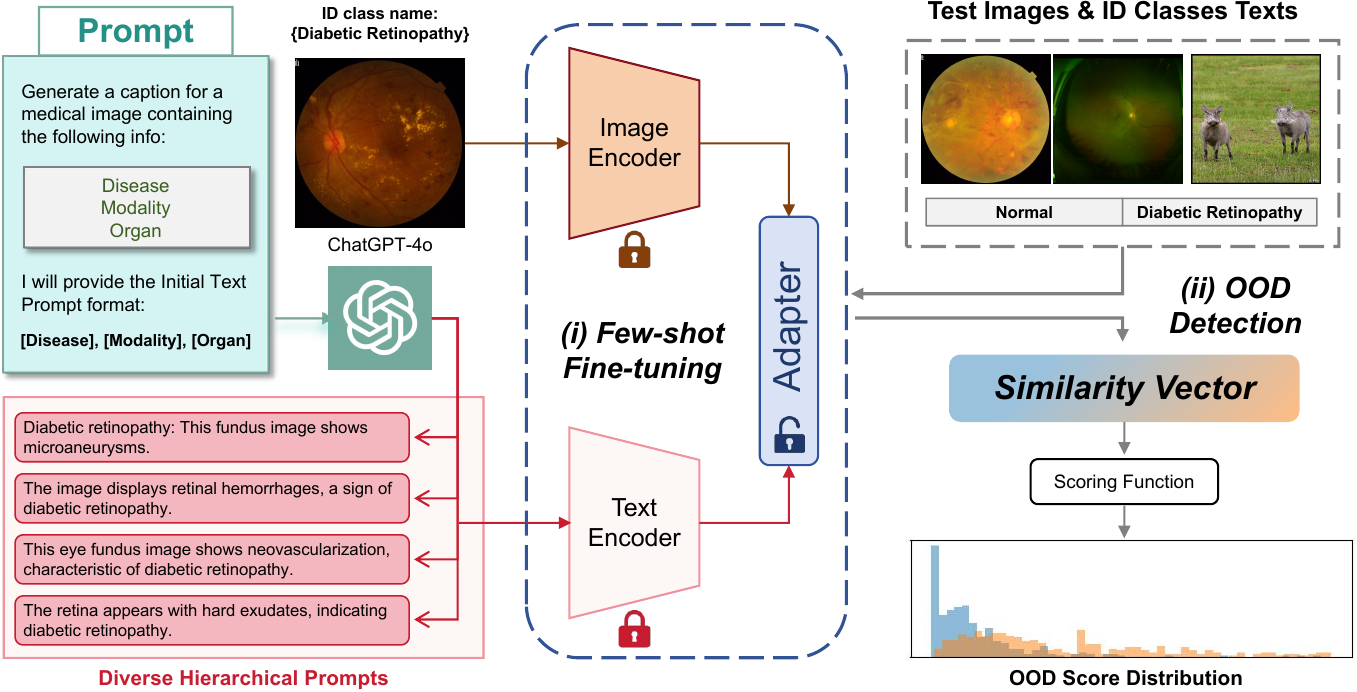}
    \caption{The fine-tuning pipeline for OOD detection with proposed hierarchical prompts.}
    \label{fig:pipeline}
\end{figure}
\subsection{Inference with Hierarchical Prompts}
Current mainstream medical VLMs commonly employ category names as textual prompts (e.g., "A {\textit{\{modality\}}} image showing {\textit{\{class name\}}}"), an intuitive approach that achieves reliable performance under standard in-distribution conditions. UniMed-CLIP~\cite{khattak2024unimed} suggests using diversifying template-captions rather than relying on any single type of prompt as input. Building on this foundation, we develop hierarchical prompts that integrate multi-level clinical semantics, such as diagnostic criteria, lesion morphology, imaging modalities, and anatomical context, to refine the discriminative boundary between ID and OOD samples. Hierarchical prompts are better aligned with medical imaging than natural images due to domain-specific characteristics: (1) Medical diagnosis relies on structured hierarchies (e.g., anatomical location → pathology severity → lesion morphology), which naturally map to multi-level prompts; (2) While natural images often lack standardized descriptors beyond generic labels (e.g., "dog"), medical images require granular, protocol-driven annotations (e.g., "non-proliferative diabetic retinopathy with microaneurysms in the superior quadrant"), enabling prompts to leverage clinically validated taxonomies.

\subsection{Few-shot Fine-tuning with Hierarchical Prompts}
Given the outstanding generalization abilities of CLIP-like models, it is also evident that fine-tuning plays a crucial role in both ID recognition and OOD detection~\cite{noda2025benchmark}. Medical CLIP-like models strive to offer a generalized feature representation by pre-training on the collected large-scale image-text pairs~\cite{khattak2024unimed}. This generalization capacity not only enables effective zero-shot transfer but also boosts the performance of few-shot fine-tuning performance when working with limited downstream data, surpassing the results of training from scratch. Fig.~\ref{fig:pipeline} presents an overview of few-shot fine-tuning pipeline, which is also in excellent alignment with our proposed hierarchical prompts~\cite{ju2023hierarchical,ju2024explore}.

\section{Experiments}
\subsection{Full-Spectrum Medical OOD Detection Benchmark}
\textbf{Medical Vision-language Models.} We include five general-purpose models (GPMs)—Meta CLIP~\cite{radford2021learning}, BioMedCLIP~\cite{zhang2023biomedclip}, PMC-CLIP~\cite{lin2023pmc}, and UniMedCLIP~\cite{khattak2024unimed}—to investigate their generalization capabilities compared to domain-specific medical VLMs. Domain-Specific Models (DSMs) refer to models originally designed for use within a single medical domain. Subsequently, we select one DSM for each medical domain, which are: FLAIR~\cite{silva2025foundation} for Ophthalmology; QuilNet~\cite{ikezogwo2023quilt} for Pathology;  and MedCLIP~\cite{wang2022medclip} for Radiology.

\begin{table}[t]

\centering
\caption{Evaluated Datasets for Full-spectrum OOD Detection Benchmark.}
\begin{tabular}{lcccccc}
\hline
             \multicolumn{4}{c|}{Foundational Dataset}                         & \multicolumn{3}{c}{Covariate Shift OOD Dataset} \\ \hline
\multicolumn{1}{l}{Dataset}        & ID Class Name         & Attribute    & \multicolumn{1}{c|}{NoI/NoO} & Dataset           & Attribute       & NoO    \\ \hline
\multicolumn{1}{l}{FIVES} & FIVES        & Normal/DR     & \multicolumn{1}{c|}{2 / 2}     & DeepDRiD       & UWF           & 2      \\
\multicolumn{1}{l}{LC25000}     & LC25000      & Benign/ACA       & \multicolumn{1}{c|}{2 / 1}     & LC25000        & Colon         & 2      \\
\multicolumn{1}{l}{COVID-19}     & COVID-19      & Normal/Pneu.      & \multicolumn{1}{c|}{2 / 1}     & COVID-19    & CT            & 2      \\ \hline

\multicolumn{1}{l}{}        & OOD Class Name            & \multicolumn{2}{c|}{ID/OOD Samples} & Class Name           & \multicolumn{2}{c}{OOD Samples}    \\ \hline
\multicolumn{1}{l}{FIVES} & AMD/Glaucoma             & \multicolumn{2}{c|}{300 / 300}     & Normal/DR       & \multicolumn{2}{c}{204}      \\
\multicolumn{1}{l}{LC25000}     & SCC             & \multicolumn{2}{c|}{300 / 300}     & Benign/ACA        & \multicolumn{2}{c}{300}     \\
\multicolumn{1}{l}{COVID-19}     & COVID-19 Pneu.            & \multicolumn{2}{c|}{300 / 300}     & Normal/Pneu.    & \multicolumn{2}{c}{300}      \\ \hline
\multicolumn{7}{l}{* NoI: Number of ID categories; NoO: Number of OOD categories.  }  \\
\multicolumn{7}{l}{* DR: diabetic retinopathy; AMD: Age-related Macular Degeneration.} \\
\multicolumn{7}{l}{* ACA: adenocarcinoma; SCC: Squamous cell carcinomas; Pneu.: Pneumonia.}
\end{tabular}
\label{Table: Benchmark dataset}
\end{table}

\noindent \textbf{Datasets.} As outlined in Table~\ref{Table: Benchmark dataset}, our benchmark leverages four foundational datasets: FIVES~\cite{jin2022fives}, ISIC 2019~\cite{tschandl2018ham10000}, LC25000~\cite{borkowski2019lung}, and COVID-19~\cite{cohen2020covid}. We first define ID categories within each dataset, then construct semantic shifted OOD samples by selecting some other classes that share imaging modalities or anatomical regions with ID data but differ in diagnostic labels. For covariate shifts, we preserve ID semantic labels while introducing distributional variations through differences in imaging devices, acquisition protocols, or population distribution. These covariate shifted OOD samples are sourced both from the foundational datasets and external repositories, such as DeepDRiD~\cite{liu2022deepdrid}. Additionally, we include 300 randomly selected ImageNet samples as far-OOD examples to simulate natural image outliers. Further details, including input prompts used and complete dataset statistics, will be released along with the codes.

\noindent \textbf{Baseline Methods and Setup. } To validate the effectiveness of the proposed hierarchical prompts, we evaluate representative zero-shot~\cite{liu2020energy,miyai2025gl,ming2022delving} and few-shot~\cite{zhou2022conditional,miyai2023locoop,zhang2021tip,xia2023hgclip} CLIP-based OOD detection methods with 50-shot fine-tuning. Performance is assessed using the area under the receiver operating characteristic curve (AUROC) for both ID recognition and OOD detection tasks.

\subsection{Main Results}
\begin{figure}[t]
    \centering
    \includegraphics[width=12cm]{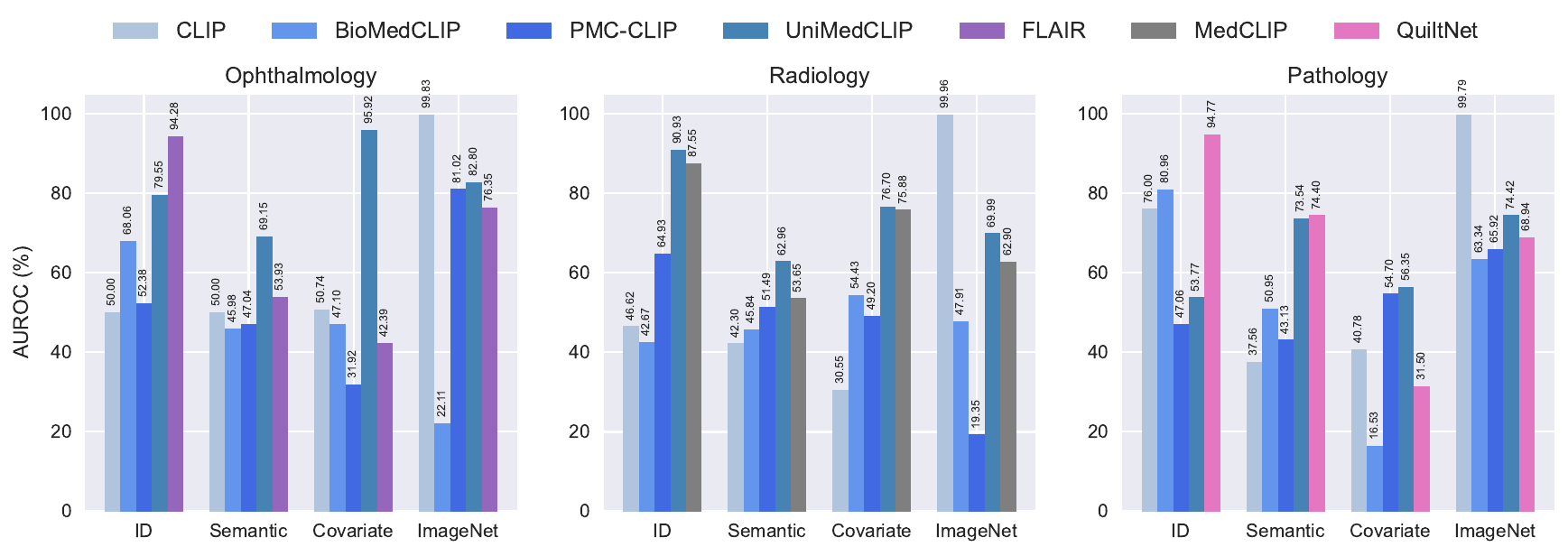}
    \caption{The comparison results across various GPMs and DSMs.}
    \label{fig:GPMS}
\end{figure}

\textbf{Comparison study across various GPMs and DSMs.} In Fig.~\ref{fig:GPMS}, we present the performance differences between GPMs and DSMs in various tasks, including ID recognition and full-spectrum OOD detection. It reveals a critical trade-off: while DSMs achieve stronger ID recognition over GPMs (e.g., 94.28\% AUC with FLAIR) through medical fine-tuning, they falter under semantic/covariate OOD shifts, exposing limitations of standard detection methods like MCM. Notably, while Meta CLIP fails to classify medical ID samples but detects far OOD (ImageNet) with 99\% AUROC, likely due to its limited medical pre-training enabling learned feature separation from natural images.

\begin{table}[t]
\addtolength{\tabcolsep}{1.7pt}
\centering
\caption{Comparison of representative CLIP-based OOD detection methods.}
\begin{tabular}{c|l|ccc|ccc|ccc}
\hline
\multicolumn{1}{l|}{}      &              & \multicolumn{3}{c|}{FLAIR} & \multicolumn{3}{c|}{UniMedCLIP} & \multicolumn{3}{c}{QuiltNet} \\ \hline
\multicolumn{1}{l|}{}      & Method       & S       & C       & I      & S         & C        & I        & S        & C       & I       \\ \hline
\multirow{6}{*}{\textit{Zero-Shot}} & Max-Logits   & 40.3    & 42.8    & 73.8   & 44.0      & 62.5     & 47.2     & 60.9     & 38.1    & 83.2    \\
                           & Energy~\cite{liu2020energy}       & 39.8    & 43.8    & 71.4   & 43.7      & 58.9     & 45.4     & 49.9     & 43.5    & \textbf{84.0}    \\
                           & GL-MCM~\cite{miyai2025gl}      & 52.1    & 43.5    & 74.1   & 54.5      & 66.3     & 54.2     & 50.3     & \textbf{44.5}    & 55.1    \\
                           & MCM~\cite{ming2022delving}          & 53.9    & 42.4    & 76.4   & 53.7      & 75.9     & \textbf{62.9}     & 55.9     & 39.8    & 51.9    \\
                           & MCM (\textit{L}=1)    & 61.6    & 65.6    & 52.2   & 67.9      & 79.5     & 13.3     & 48.6     & 40.3    & 69.5    \\
                           & MCM (\textit{L}=5)    & \textbf{66.7}    & \textbf{87.7}    & \textbf{82.4}   & \textbf{71.1}      & \textbf{82.5}     & 35.1     & \textbf{63.0}     & 37.5    & 59.3    \\ \hline
\multirow{5}{*}{\textit{Few-Shot}}  & CoOp~\cite{zhou2022conditional}         & 70.3    & 45.6    & 90.4   & 76.4      & \textbf{83.5}     & 56.2     & 67.2     & 55.6    & 66.4    \\
                           & LoCoOp~\cite{miyai2023locoop}       & 72.6    & 52.3    & \textbf{92.1}   & 74.5      & 71.2     & 45.7     & 55.6     & \textbf{57.2}    & 63.1    \\
                           & TipAdapter~\cite{zhang2021tip}  & 68.9    & 50.6    & 80.4   & 72.5      & 80.4     & \textbf{64.5}     & 61.3     & 40.1    & 50.0    \\
                           & HGCLIP~\cite{xia2023hgclip}       & 54.5    & 43.5    & 56.5   & 68.3      & 55.4     & 50.8     & 50.4     & 48.2    & 55.7    \\
                           & LoCoOp (\textit{L}=5) & \textbf{74.1}    & \textbf{62.4}    & 88.3   & \textbf{77.9}      & 82.9     & 44.2     & \textbf{69.5}     & 49.9    & \textbf{66.7}    \\ \hline
\multicolumn{11}{l}{* S: Semantic shifts; C: Covariate shifts; I: ImageNet;.} 
\end{tabular}
\label{comparison study}
\end{table}

\noindent \textbf{Comparison study on advanced methods.} We select the models that perform best in ID recognition in each medical domain in Fig.~\ref{fig:GPMS}, to further examine the performance of advanced CLIP-based OOD detection methods. The results are shown in Table~\ref{comparison study}. It is found that no single method demonstrates universal superiority across all types of OOD scenarios. Specifically, MCM with $L=5$ shows an improvement in detecting covariate shifts (e.g., FLAIR achieves 87.7\% AUROC) but severely degrades on ImageNet for the other two models. We observe that some medical VLMs tend to classify natural images all into a single category, resulting in the overconfident predictions with catastrophic results (e.g., UniMedCLIP on MCM (L=1) with 13.1\% AUROC). This suggests that there should be further potential to improve the scoring functions. Few-shot fine-tuning serves as a universal enhancer for almost all baselines. Relying on the generalization ability of the CLIP-based models, few-shot fine-tuning with ID samples can help focus more on the categories of interest. Meanwhile, hierarchical prompts enrich the models with fine-grained semantic information, thereby enhancing the robustness of the models at critical decision boundaries.

\begin{figure}[t]
    \centering
    \includegraphics[width=12cm]{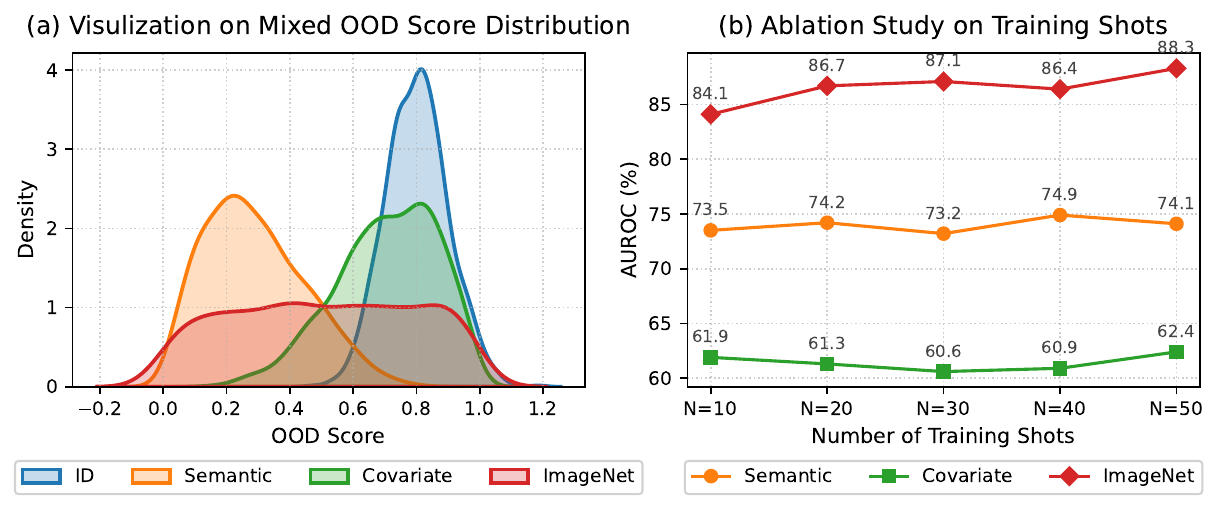}
    \caption{(a) The visualization on mixed OOD score distribution. (b) Few-shot OOD detection results with different numbers of ID training samples for fine-tuning.}
    \label{fig:enter-label}
\end{figure}

\subsection{Analysis}
\textbf{OOD Detection Performance with Mixed Types of Shift.} 
For more practical deployment scenarios, a robust medical VLM must be capable of identifying OOD samples from different types of distribution shifts and reacting accordingly. For instance, when encountering OOD samples with covariate shifts, the system could redirect them to specialized expert models for appropriate ID recognition. In this context, we visualize the score distribution density of FLAIR+LoCoOp ($L=5$) with different OOD types. As shown in Fig.~\ref{fig:enter-label}-(a), covariate shifts pose a greater challenge due to their higher semantic similarity with ID samples. Meanwhile, the distribution of ImageNet OOD overlaps significantly with that of the other two OOD types. Such misclassifications could potentially trigger unnecessary diagnostic procedures by distributed specialized medical expert models.

\noindent \textbf{Impact of Training Sample Size.} Fig.~\ref{fig:enter-label}-(b) presents the results with different numbers of ID training shots for FLAIR+LoCoOp ($L=5$). Our analysis reveals that LoCoOp demonstrates low sensitivity to the training sample size, achieving robust performance even with minimal ID data. This capability stems from the strong generalization inherent in pre-trained medical VLMs. Notably, while incorporating additional training samples yields marginal performance gains, the model maintains stable efficacy across all tested configurations.

\section{Conclusion}
This work establishes the first comprehensive benchmark for medical OOD detection with VLMs, evaluating both general-purpose and domain-specific CLIP-like models under full-spectrum shifts. The proposed hierarchical prompt-based method significantly enhances OOD separability for medical VLMs by leveraging structured medical semantics. We hope that this benchmark and its findings will inspire further research to address critical challenges in VLM-based OOD detection, ultimately contributing to the development of trustworthy and reliable medical diagnostic systems for real-world applications.
% \section{Acknowledgments}
% Dr. Keane is supported by a UK Research \& Innovation Future Leaders Fellowship (MR/T019050/1) and The Rubin Foundation Charitable Trust.

%
\bibliographystyle{splncs04}
\bibliography{mybibliography}

\begin{thebibliography}{10}
\providecommand{\url}[1]{\texttt{#1}}
\providecommand{\urlprefix}{URL }
\providecommand{\doi}[1]{https://doi.org/#1}

\bibitem{borkowski2019lung}
Borkowski, A.A., Bui, M.M., Thomas, L.B., Wilson, C.P., DeLand, L.A., Mastorides, S.M.: Lung and colon cancer histopathological image dataset (lc25000). arXiv preprint arXiv:1912.12142  (2019)

\bibitem{cohen2020covid}
Cohen, J.P., Morrison, P., Dao, L.: Covid-19 image data collection. arXiv preprint arXiv:2003.11597  (2020)

\bibitem{hong2024out}
Hong, Z., Yue, Y., Chen, Y., Cong, L., Lin, H., Luo, Y., Wang, M.H., Wang, W., Xu, J., Yang, X., et~al.: Out-of-distribution detection in medical image analysis: A survey. arXiv preprint arXiv:2404.18279  (2024)

\bibitem{ikezogwo2023quilt}
Ikezogwo, W., Seyfioglu, S., Ghezloo, F., Geva, D., Sheikh~Mohammed, F., Anand, P.K., Krishna, R., Shapiro, L.: Quilt-1m: One million image-text pairs for histopathology. Advances in neural information processing systems  \textbf{36},  37995--38017 (2023)

\bibitem{jin2022fives}
Jin, K., Huang, X., Zhou, J., Li, Y., Yan, Y., Sun, Y., Zhang, Q., Wang, Y., Ye, J.: Fives: A fundus image dataset for artificial intelligence based vessel segmentation. Scientific data  \textbf{9}(1), ~475 (2022)

\bibitem{ju2023hierarchical}
Ju, L., Yu, Z., Wang, L., Zhao, X., Wang, X., Bonnington, P., Ge, Z.: Hierarchical knowledge guided learning for real-world retinal disease recognition. IEEE Transactions on Medical Imaging  \textbf{43}(1),  335--350 (2023)

\bibitem{ju2024explore}
Ju, L., Zhou, Y., Xia, P., Alexander, D., Keane, P.A., Ge, Z.: Explore vision-language model with hierarchical information for multiple retinal disease recognition. Investigative Ophthalmology \& Visual Science  \textbf{65}(7),  1593--1593 (2024)

\bibitem{khattak2024unimed}
Khattak, M.U., Kunhimon, S., Naseer, M., Khan, S., Khan, F.S.: Unimed-clip: Towards a unified image-text pretraining paradigm for diverse medical imaging modalities. arXiv preprint arXiv:2412.10372  (2024)

\bibitem{lee2018simple}
Lee, K., Lee, K., Lee, H., Shin, J.: A simple unified framework for detecting out-of-distribution samples and adversarial attacks. Advances in neural information processing systems  \textbf{31} (2018)

\bibitem{liang2018enhancing}
Liang, S., Li, Y., Srikant, R.: Enhancing the reliability of out-of-distribution image detection in neural networks. In: International Conference on Learning Representations (2018)

\bibitem{lin2023pmc}
Lin, W., Zhao, Z., Zhang, X., Wu, C., Zhang, Y., Wang, Y., Xie, W.: Pmc-clip: Contrastive language-image pre-training using biomedical documents. In: International Conference on Medical Image Computing and Computer-Assisted Intervention. pp. 525--536. Springer (2023)

\bibitem{liu2022deepdrid}
Liu, R., Wang, X., Wu, Q., Dai, L., Fang, X., Yan, T., Son, J., Tang, S., Li, J., Gao, Z., et~al.: Deepdrid: Diabetic retinopathy—grading and image quality estimation challenge. Patterns  \textbf{3}(6) (2022)

\bibitem{liu2020energy}
Liu, W., Wang, X., Owens, J., Li, Y.: Energy-based out-of-distribution detection. Advances in neural information processing systems  \textbf{33},  21464--21475 (2020)

\bibitem{ming2022delving}
Ming, Y., Cai, Z., Gu, J., Sun, Y., Li, W., Li, Y.: Delving into out-of-distribution detection with vision-language representations. Advances in neural information processing systems  \textbf{35},  35087--35102 (2022)

\bibitem{ming2024does}
Ming, Y., Li, Y.: How does fine-tuning impact out-of-distribution detection for vision-language models? International Journal of Computer Vision  \textbf{132}(2),  596--609 (2024)

\bibitem{miyai2023locoop}
Miyai, A., Yu, Q., Irie, G., Aizawa, K.: Locoop: Few-shot out-of-distribution detection via prompt learning. Advances in Neural Information Processing Systems  \textbf{36},  76298--76310 (2023)

\bibitem{miyai2023zero}
Miyai, A., Yu, Q., Irie, G., Aizawa, K.: Zero-shot in-distribution detection in multi-object settings using vision-language foundation models. arXiv preprint arXiv:2304.04521  (2023)

\bibitem{miyai2025gl}
Miyai, A., Yu, Q., Irie, G., Aizawa, K.: Gl-mcm: Global and local maximum concept matching for zero-shot out-of-distribution detection. International Journal of Computer Vision pp. 1--11 (2025)

\bibitem{noda2025benchmark}
Noda, S., Miyai, A., Yu, Q., Irie, G., Aizawa, K.: A benchmark and evaluation for real-world out-of-distribution detection using vision-language models. arXiv preprint arXiv:2501.18463  (2025)

\bibitem{radford2021learning}
Radford, A., Kim, J.W., Hallacy, C., Ramesh, A., Goh, G., Agarwal, S., Sastry, G., Askell, A., Mishkin, P., Clark, J., et~al.: Learning transferable visual models from natural language supervision. In: International conference on machine learning. pp. 8748--8763. PmLR (2021)

\bibitem{silva2025foundation}
Silva-Rodriguez, J., Chakor, H., Kobbi, R., Dolz, J., Ayed, I.B.: A foundation language-image model of the retina (flair): Encoding expert knowledge in text supervision. Medical Image Analysis  \textbf{99},  103357 (2025)

\bibitem{tschandl2018ham10000}
Tschandl, P., Rosendahl, C., Kittler, H.: The ham10000 dataset, a large collection of multi-source dermatoscopic images of common pigmented skin lesions. Scientific data  \textbf{5}(1), ~1--9 (2018)

\bibitem{wang2024common}
Wang, M., Lin, T., Lin, A., Yu, K., Peng, Y., Wang, L., Chen, C., Zou, K., Liang, H., Chen, M., et~al.: Common and rare fundus diseases identification using vision-language foundation model with knowledge of over 400 diseases. arXiv preprint arXiv:2406.09317  (2024)

\bibitem{wang2022medclip}
Wang, Z., Wu, Z., Agarwal, D., Sun, J.: Medclip: Contrastive learning from unpaired medical images and text. In: Proceedings of the Conference on Empirical Methods in Natural Language Processing. Conference on Empirical Methods in Natural Language Processing. vol.~2022, p.~3876 (2022)

\bibitem{wu2024mm}
Wu, R., Zhang, C., Zhang, J., Zhou, Y., Zhou, T., Fu, H.: Mm-retinal: Knowledge-enhanced foundational pretraining with fundus image-text expertise. In: International Conference on Medical Image Computing and Computer-Assisted Intervention. pp. 722--732. Springer (2024)

\bibitem{xia2023hgclip}
Xia, P., Yu, X., Hu, M., Ju, L., Wang, Z., Duan, P., Ge, Z.: Hgclip: exploring vision-language models with graph representations for hierarchical understanding. arXiv preprint arXiv:2311.14064  (2023)

\bibitem{yang2023full}
Yang, J., Zhou, K., Liu, Z.: Full-spectrum out-of-distribution detection. International Journal of Computer Vision  \textbf{131}(10),  2607--2622 (2023)

\bibitem{zhang2021tip}
Zhang, R., Fang, R., Zhang, W., Gao, P., Li, K., Dai, J., Qiao, Y., Li, H.: Tip-adapter: Training-free clip-adapter for better vision-language modeling. arXiv preprint arXiv:2111.03930  (2021)

\bibitem{zhang2023biomedclip}
Zhang, S., Xu, Y., Usuyama, N., Xu, H., Bagga, J., Tinn, R., Preston, S., Rao, R., Wei, M., Valluri, N., et~al.: Biomedclip: a multimodal biomedical foundation model pretrained from fifteen million scientific image-text pairs. arXiv preprint arXiv:2303.00915  (2023)

\bibitem{zhou2022conditional}
Zhou, K., Yang, J., Loy, C.C., Liu, Z.: Conditional prompt learning for vision-language models. In: Proceedings of the IEEE/CVF conference on computer vision and pattern recognition. pp. 16816--16825 (2022)

\end{thebibliography}

\end{document}